\documentclass[11pt,a4paper]{article}

\usepackage[margin=2.2cm,headheight=14pt,headsep=8mm,footskip=10mm]{geometry}
\usepackage{iftex}
\ifPDFTeX
  \usepackage[T1]{fontenc}
\fi
\usepackage{newtxtext,newtxmath}
\usepackage[table]{xcolor}
\usepackage{graphicx,booktabs,tabularx,array,amsmath}
\usepackage{microtype,setspace,titlesec,fancyhdr,enumitem}
\usepackage[most]{tcolorbox}
\usepackage{tikz}
\usetikzlibrary{arrows.meta,positioning,fit,calc}
\usepackage{placeins,needspace,indentfirst,cite,hyperref,caption,float}

\definecolor{darkblue}{HTML}{17335D}
\definecolor{reportblue}{HTML}{2766A3}
\definecolor{paleBlue}{HTML}{EEF3F8}
\definecolor{paleGray}{HTML}{F5F6F7}
\definecolor{midGray}{HTML}{697581}
\definecolor{linegray}{HTML}{9AA5B1}
\definecolor{warmAmber}{HTML}{A97943}
\definecolor{deepGreen}{HTML}{557965}

\hypersetup{colorlinks=true,linkcolor=darkblue,citecolor=reportblue,urlcolor=reportblue,pdftitle={Compounding Physical RSI: Turning Execution Failure into Inherited Capability},pdfauthor={CSU MinosLab}}
\titleformat{\section}{\large\bfseries\color{darkblue}}{\thesection}{0.7em}{}
\titleformat{\subsection}{\normalsize\bfseries\color{darkblue}}{\thesubsection}{0.6em}{}
\titlespacing*{\section}{0pt}{8pt plus 1pt minus 1pt}{3pt}
\titlespacing*{\subsection}{0pt}{5pt plus 1pt minus 1pt}{2pt}
\setlist[itemize]{leftmargin=1.7em,itemsep=2pt,topsep=3pt}
\newcolumntype{Y}{>{\raggedright\arraybackslash}X}
\newtcolorbox{definitionbox}{enhanced,colback=paleBlue,colframe=reportblue,boxrule=0pt,
  leftrule=2pt,arc=0pt,left=3mm,right=3mm,top=1.6mm,bottom=1.6mm,before skip=6pt,after skip=6pt}

\begin{document}
\thispagestyle{fancy}

\begin{center}
{\color{reportblue}\rule{\textwidth}{0.65pt}}\par\vspace{0.06cm}
{\fontsize{15.5}{18.5}\selectfont\bfseries\color{darkblue}
Compounding Physical RSI:\\
Turning Execution Failure into Inherited Capability\par}
\vspace{0.03cm}
{\fontsize{10}{12}\selectfont Technical Report\par}
\vspace{0.05cm}
{\color{reportblue}\rule{\textwidth}{0.65pt}}
\end{center}
\vspace{-0.05cm}

\begingroup
\fontsize{9.2}{11.0}\selectfont
\begin{center}
Yixuan Yang$^{1}$,\enspace Qiang Chen$^{1}$,\enspace Run Shao$^{1}$,\enspace Shuwei Dong$^{1}$,\enspace Wenhao Li$^{1}$,\enspace Yunpeng Wang$^{1}$\\[0.5pt]
Wenhan Fang$^{1}$,\enspace Haomiao Wu$^{1}$,\enspace Xingyu Wang$^{1}$,\enspace Lichen Man,\enspace Rong Zhang\\[0.5pt]
Xuezhi Cui,\enspace Yuan Liu,\enspace Bingyan Lu,\enspace Jiawei Ma,\enspace Xiaoyang Ma,\enspace Zhemin Sun\\[0.5pt]
Zhuorui Yao,\enspace Yutong Dong,\enspace Yuxuan Zhao,\enspace Yichong Mu,\enspace Jiabao Jiang\\[0.5pt]
Wenxuan Deng,\enspace Junling Zhou,\enspace Yi Chen$^{2}$,\enspace Deyu Zhang$^{3}$,\enspace Yiming Yan$^{3}$\\[0.5pt]
Yige Yuan$^{3}$,\enspace Shan You,\enspace Fei Wang$^{2}$,\enspace Haifeng Li$^{4}$,\enspace Hongyan Xu$^{2}$,\enspace Yichao Cao$^{5}$,\enspace Xiu Su$^{6,*}$
\end{center}
\endgroup

\vspace{-0.02cm}
\begin{center}
{\fontsize{8.6}{10.0}\selectfont\bfseries Author Contributions\par}\vspace{0.02cm}
{\fontsize{8.2}{10.0}\selectfont
\makebox[\textwidth][c]{%
\begin{minipage}{0.76\textwidth}
\begin{tabular}{@{}p{0.47\linewidth}@{\hspace{0.06\linewidth}}p{0.47\linewidth}@{}}
$^{1}$ Core Algorithm Development \& Implementation & $^{2}$ Principal Contributor\\[1pt]
$^{3}$ Resource Contributor & $^{4}$ Project Leader\\[1pt]
$^{5}$ Execution Leader \& Principal Contributor & $^{6}$ Technical Leader \& Principal Contributor
\end{tabular}
\end{minipage}}\\[0.5pt]
$^{*}$ Corresponding author\par}
\end{center}
\vspace{-0.02cm}
\begin{center}
{\fontsize{9.4}{11.0}\selectfont\bfseries CSU MinosLab\par}\vspace{0.01cm}
{\fontsize{9.0}{10.5}\selectfont ECCV 2026 RoboDojo Safe World Models Challenge
\quad | \quad
Champion
\quad | \quad
CSU-AI-07\par}
\end{center}
\vspace{0.01cm}

\begin{tcolorbox}[
  enhanced,colback=white,colframe=reportblue,boxrule=0.75pt,arc=2mm,
  left=3.2mm,right=3.2mm,top=1.6mm,bottom=1.7mm,before skip=0pt,after skip=0pt]
\setstretch{1.02}
\begin{center}
{\color{reportblue}\fontsize{13}{15}\selectfont\bfseries Abstract}
\end{center}
\vspace{-0.12cm}
\fontsize{9.8}{11.8}\selectfont
\setlength{\parindent}{1.5em}
\setlength{\parskip}{2.0pt}
\noindent\hspace*{1.5em}Generalist robot policies are predominantly trained on static offline datasets, yet physical manipulation unfolds sequentially across stages. Because a weakness in a downstream stage remains unobservable until the policy reliably passes earlier ones—a robot that fails to grasp an object never reveals how it would fail during transport—each resolved bottleneck exposes the next hidden failure mode. This staged structure calls for Recursive Self-Improvement (RSI), in which a policy iteratively executes, diagnoses its own failures, and updates itself. Moreover, because overall task completion behaves like a product of stage-level pass rates, recursive gains can multiply across generations if each round builds upon the last. In physical environments, however, three bottlenecks obstruct compounding: terminal execution feedback rarely localizes the stage that caused a failure, failed trajectories do not supply the corrective action, and updating shared parameters to fix one failure mode can degrade previously mastered skills.

We propose \textbf{Compounding Physical RSI}, a self-evolving policy framework that turns execution failures into heritable capability across generations. First, to localize failure causes, automated capability-gap assessment traces failed rollouts to the exact task stage where progression breaks down and formulates a structured improvement objective. Second, conditioned directly on this objective, our remedy stage curates validated, stage-relevant experience and refines the policy via behavior-regularized offline reinforcement learning, achieving targeted improvement without requiring costly online rollouts at each optimization step. Third, to safeguard existing skills, our retention stage constructs weight-space candidates between the inherited reference policy and the refined policy, evaluates them across the full task suite, and promotes the best-performing candidate as the reference state for the next generation.

Evaluated across all 42 tasks of the RoboDojo-Sim benchmark, our resulting policy, \textbf{CSU-AI-07}, ranks first on the overall leaderboard with a three-seed mean Overall Score of 26.15 and Success Rate of 20.70\%, outperforming the next-best entry by 5.92 points (a 29\% relative gain) and 5.82 percentage points (a 39\% relative gain).

\vspace{1.5pt}
\noindent\textbf{Keywords:} Compounding Physical RSI, Recursive Self-Improvement, Vision-Language-Action, capability inheritance, offline reinforcement learning, RoboDojo-Sim
\end{tcolorbox}

\clearpage
\section{Introduction}
Vision-language-action (VLA) models have substantially advanced generalist robot manipulation by unifying visual perception, language grounding, and action generation within a single policy \cite{rt1,rt2,openvla,pi0,pi05}. Recent efforts have extended this paradigm across diverse embodiments \cite{openx,octo,robocat}, refined spatial representations and fine-tuning efficiency \cite{spatialforcing,openvlaoft}, and scaled pre-training to over 100K hours of real-world trajectories \cite{xiaomirobotics1}. Yet despite these advances in pre-training scale, a generalist policy still encounters many of its fundamental capability limits only after static training has ended and closed-loop physical execution begins.

This limitation stems from the sequential, stage-gated nature of manipulation. A robot must approach an object before grasping it, and must establish a stable grasp before transporting or placing it. Consequently, a deficiency in a downstream stage remains masked until the policy reliably passes all preceding stages: a policy that fails during grasping never reveals how it would fail during transport, because its rollouts never reach that state. Moreover, the failure modes encountered during transport depend directly on the contact configuration produced by the grasp, as each policy induces its own closed-loop state distribution \cite{dagger}. We define the earliest task stage that a policy cannot yet pass reliably as its \emph{capability frontier}. Because behavior beyond this frontier remains unobserved during execution, a fixed dataset cannot anticipate how the evolving policy will fail once it crosses the current barrier. Each successful fix advances the capability frontier and exposes the next latent failure mode.

A moving capability frontier naturally motivates Recursive Self-Improvement (RSI) for embodied intelligence. Rooted in classical formulations of self-referential learning \cite{good,schmidhuber} and renewed in modern foundation-model research \cite{duan}, RSI recognizes that when a policy's next bottleneck becomes visible only after the current one is resolved, improvement cannot be reduced to a single static training run. Instead, the robot must iteratively execute, diagnose its own failures, learn from them, and execute again so that each generation uncovers the empirical evidence required by the next. Physical RSI formalizes this paradigm for embodied systems, where contact-rich physical experience drives persistent policy evolution \cite{whoimproves,towardrsi}. Recent studies have explored key facets of embodied self-improvement, including inference-time self-correction and subtask backtracking \cite{cyclevla}, deployment-time adaptation and memory consolidation \cite{selfadaptive,mebrain}, and iterative policy updates via autonomous data collection or residual reinforcement learning \cite{robocat,pistar,xiao,embodirsi}.

While these efforts show that a robot policy can improve over a single update or within narrow task settings, the central open question is whether such improvements can \emph{compound} across generations. The potential for compounding follows directly from the staged structure of manipulation. Because a task succeeds only when every constituent stage succeeds, overall task completion behaves like a product of stage-level pass rates. In such a product, stage-level gains multiply: stabilizing a grasp not only lifts the completion rate of every subsequent stage, but also sends a larger fraction of rollouts across the capability frontier, generating richer downstream evidence for the next generation. A recursive loop can therefore achieve super-additive progress across rounds, provided that each generation builds reliably upon the verified gains of its predecessor.

However, making Physical RSI compound in practice is hindered by three fundamental bottlenecks that mirror the lifecycle of a policy update. First, execution feedback suffers from opaque temporal credit assignment \cite{suttonbarto}: a terminal failure flag indicates that a trial failed, but rarely localizes which stage broke, as visible failure often manifests long after the erroneous action that caused it. Second, failed rollouts lack corrective supervision: even a precisely localized failure shows what went wrong without specifying the counterfactual action the policy should have executed. Third, multi-task gains are fragile under shared weights: because a generalist policy encodes all skills in one parameter set, an unconstrained update that repairs one failure stage can silently degrade previously mastered tasks \cite{ewc}. Under the same multiplicative structure that amplifies stage-level gains, such cross-task regressions multiply losses, causing naive recursive loops to accumulate policy churn rather than lasting capability.

We argue that the core challenge of Physical RSI is therefore achieving \emph{compounding}—ensuring that every generation diagnoses its active frontier, resolves it effectively, and inherits verified capabilities without regression. To this end, we propose \textbf{Compounding Physical RSI}, a three-stage self-evolving policy framework, and evaluate the resulting generalist policy, \textbf{CSU-AI-07}, across all 42 tasks of RoboDojo-Sim. Our main contributions are:
\begin{itemize}
  \item \textbf{Formulating compounding as the central problem of Physical RSI.} We show how the stage-gated structure of manipulation necessitates recursive improvement and establish a formal structural definition of compounding across policy generations.
  \item \textbf{A closed-loop Diagnosis--Remedy--Retention framework.} We couple stage-level capability-gap diagnosis with objective-conditioned offline reinforcement learning (remedy) and evaluation-guided weight-space consolidation (retention), ensuring that targeted frontier fixes are absorbed without eroding prior skills.
  \item \textbf{State-of-the-art generalist benchmark performance.} Across the 42 tasks of RoboDojo-Sim, CSU-AI-07 ranks first on the overall leaderboard in both Overall Score (26.15) and Success Rate (20.70\%), outperforming thirteen public generalist policies by substantial margins.
\end{itemize}

\section{Compounding Physical RSI}
\subsection{From improvement to compounding}
The stage-gated structure of manipulation imposes three strict requirements on a recursive improvement loop. First, because downstream weaknesses remain hidden behind earlier bottlenecks, improvement is most consequential where execution currently stalls; the loop must therefore accurately localize the active capability frontier. Second, because failed rollouts carry no explicit corrective labels, the loop must synthesize targeted supervision to cross that frontier. Third, because stage-level regressions multiply just as readily as stage-level gains, the loop must preserve previously mastered capabilities when absorbing new updates. Checking these three requirements across successive generations yields a structural definition of compounding:

\begin{definitionbox}
\textbf{\color{darkblue}Compounding Physical RSI.} A recursive improvement loop is \emph{compounding} when each generation is diagnosed from its own closed-loop execution, refined specifically against that diagnosis, and promoted only as a behaviorally verified state that the next generation can inherit and build upon.
\end{definitionbox}

The three clauses of this definition map one-to-one to the three stages of our framework: \emph{Diagnosis} (Section~\ref{sec:diagnosis}), \emph{Remedy} (Section~\ref{sec:remedy}), and \emph{Retention} (Section~\ref{sec:retention}). Consistent with recent formulations of Physical RSI \cite{whoimproves}, the term \emph{physical} denotes that the loop is driven by closed-loop, contact-rich execution governed by physical dynamics—whether in physics simulation or on physical hardware. In our instantiation, closed-loop execution in RoboDojo-Sim supplies the empirical failure evidence, and full-benchmark behavioral outcomes determine which candidate policy survives into the next generation.

\subsection{The recursive policy transition}
For capability to accumulate across generations, the handover from one generation to the next must be governed by explicit operators. We formalize generation $k$ as a structured policy transition in which each intermediate product fulfills a distinct functional role. Let $\pi_k$ denote the inherited reference policy at generation $k$. Rolling out $\pi_k$ across the task suite yields execution evidence $E_k=\{(o_t,s_t,a_t,r_t,\tau_t)\}_{t=1}^{T}$, recording observations $o_t$, robot and object states $s_t$, actions $a_t$, rewards $r_t$, and task-stage annotations $\tau_t$.

\begin{figure}[H]
\centering
\includegraphics[width=\textwidth]{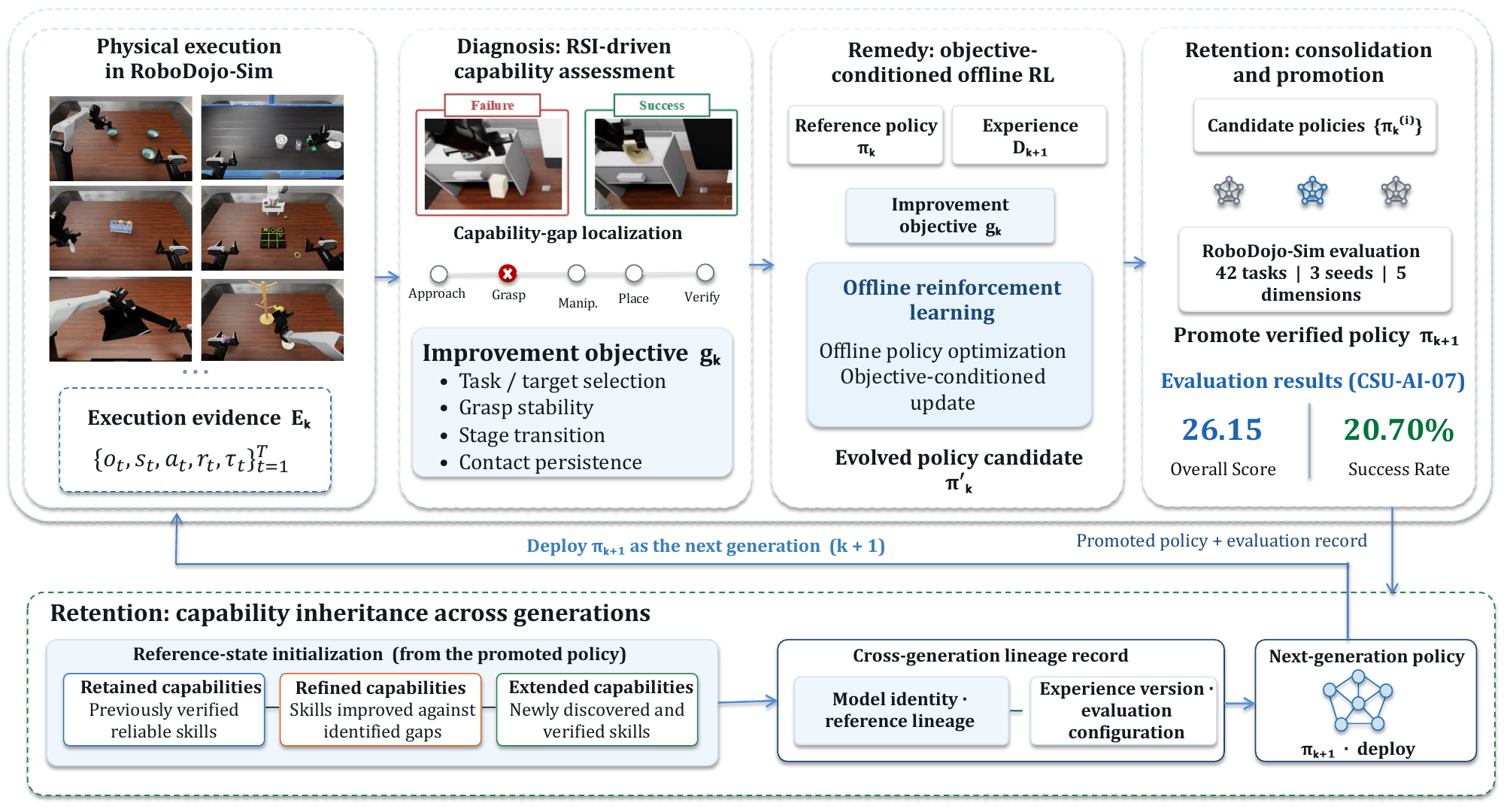}
\caption{Architecture of Compounding Physical RSI. Closed-loop execution in RoboDojo-Sim feeds automated stage-level capability-gap assessment (Diagnosis), which produces a structured objective $g_k$ to guide experience curation and offline reinforcement learning (Remedy). Retention interpolates candidates between the inherited reference policy and the refined policy, promotes the best-performing candidate via full-benchmark evaluation, and passes it forward as the reference state for the next generation.}
\label{fig:rsi_overview}
\end{figure}

\noindent One recursive generation then proceeds as:
\begin{equation}
\begin{aligned}
g_k&=\mathcal{A}(E_k), & \mathcal{B}_{k+1}&=\mathcal{B}_k\cup\mathcal{F}(E_k), & \mathcal{D}_{k+1}&=\mathcal{C}(\mathcal{B}_{k+1},g_k),\\
\pi'_k&=\mathcal{U}_{\mathrm{off}}(\pi_k,\mathcal{D}_{k+1},g_k), & \pi_{k+1}&=\mathcal{S}\!\left(\{\pi_k^{(i)}\},\mathcal{V}_{k+1}\right),
\end{aligned}
\label{eq:transition}
\end{equation}
where $\mathcal{A}$ performs stage-level capability-gap assessment to extract the structured improvement objective $g_k$; $\mathcal{F}$ validates and normalizes new execution traces before integrating them into the accumulated experience base $\mathcal{B}_k$; $\mathcal{C}$ curates the objective-relevant experience subset $\mathcal{D}_{k+1}$; $\mathcal{U}_{\mathrm{off}}$ conducts behavior-regularized offline policy refinement to produce the evolved endpoint $\pi'_k$; and $\mathcal{S}$ selects the next generation's reference policy $\pi_{k+1}$ from weight-consolidated candidates $\{\pi_k^{(i)}\}$ via automated full-benchmark evaluation $\mathcal{V}_{k+1}$. As shown in Figure~\ref{fig:rsi_overview}, the upper pathway executes this within-generation progression from rollout diagnosis to candidate promotion, while the lower pathway transfers the promoted policy $\pi_{k+1}$ forward as the inherited reference state of generation $k+1$.

This transition architecture directly addresses the practical constraints of learning from physical interaction. Because physical rollouts are costly, execution is reserved for gathering diagnostic evidence and verifying candidate policies, whereas policy optimization is performed offline over accumulated experience without requiring new environment interactions at each gradient step \cite{levine}. Equally important, because proxy training losses correlate poorly with closed-loop success in contact-rich manipulation, the recursive loop is closed by empirical task evaluation $\mathcal{V}_{k+1}$, ensuring that a candidate advances only when its complete behavioral profile justifies promotion.

\section{Diagnosis: Making Failure Legible}
\label{sec:diagnosis}
Diagnosis initiates each generation of the RSI loop by translating raw execution traces into an explicit specification of what the policy must improve next—fulfilling the first prerequisite for compounding.

\subsection{Execution evidence}
A binary terminal reward reports only whether an entire chain of manipulation stages succeeded, obscuring where a failed rollout broke down. Two policies with identical overall success rates may stall at entirely different stages and require fundamentally different updates. To preserve the temporal structure of every trial, we execute the current policy $\pi_k$ across diverse task configurations in RoboDojo-Sim, which provides a repeatable testbed for behavioral probing. Each rollout is logged as a time-indexed trajectory capturing task-stage boundaries, action transitions, robot and object state changes, terminal conditions, and behavioral responses to environmental variation.

\subsection{Stage-level capability-gap assessment}
Because manipulation progress is stage-gated, the most informative diagnostic signal in a failed rollout is the exact stage where progression halts. Downstream stages beyond that point remain unvisited—leaving little empirical basis for optimizing them—whereas resolving the bottleneck at the active capability frontier immediately benefits every subsequent stage. Our capability-gap assessment operator $\mathcal{A}$ therefore evaluates each rollout against the expected stage progression of the task (e.g., \emph{approach}, \emph{grasp}, \emph{manipulate}, \emph{place}, and \emph{verify}). By detecting where closed-loop execution diverges from the expected progression and characterizing the conditions accompanying that divergence, the assessment isolates recurring failure patterns such as target misselection, unstable grasp formation, incomplete manipulation stages, inconsistent action transitions, and sensitivity to scene variations. Operating at the granularity of task stages allows $\mathcal{A}$ to pinpoint the exact interaction segment associated with the bottleneck.

\subsection{From capability gap to improvement objective}
For diagnostic findings to steer downstream learning, they must be represented in a form that directly parameterizes data curation and policy optimization. We therefore compile the identified capability gap into a structured improvement objective $g_k$ that specifies the failing task context, the bottleneck stage, and the affected capability dimension—such as target selection, grasp stability, stage transition, or contact persistence. This structured objective $g_k$ formally marks the capability frontier of generation $k$, conditioning both experience selection ($\mathcal{C}$) and offline policy optimization ($\mathcal{U}_{\mathrm{off}}$) in the Remedy stage.

\begin{figure}[t]
\centering
\includegraphics[width=0.90\textwidth,height=0.37\textheight,keepaspectratio]{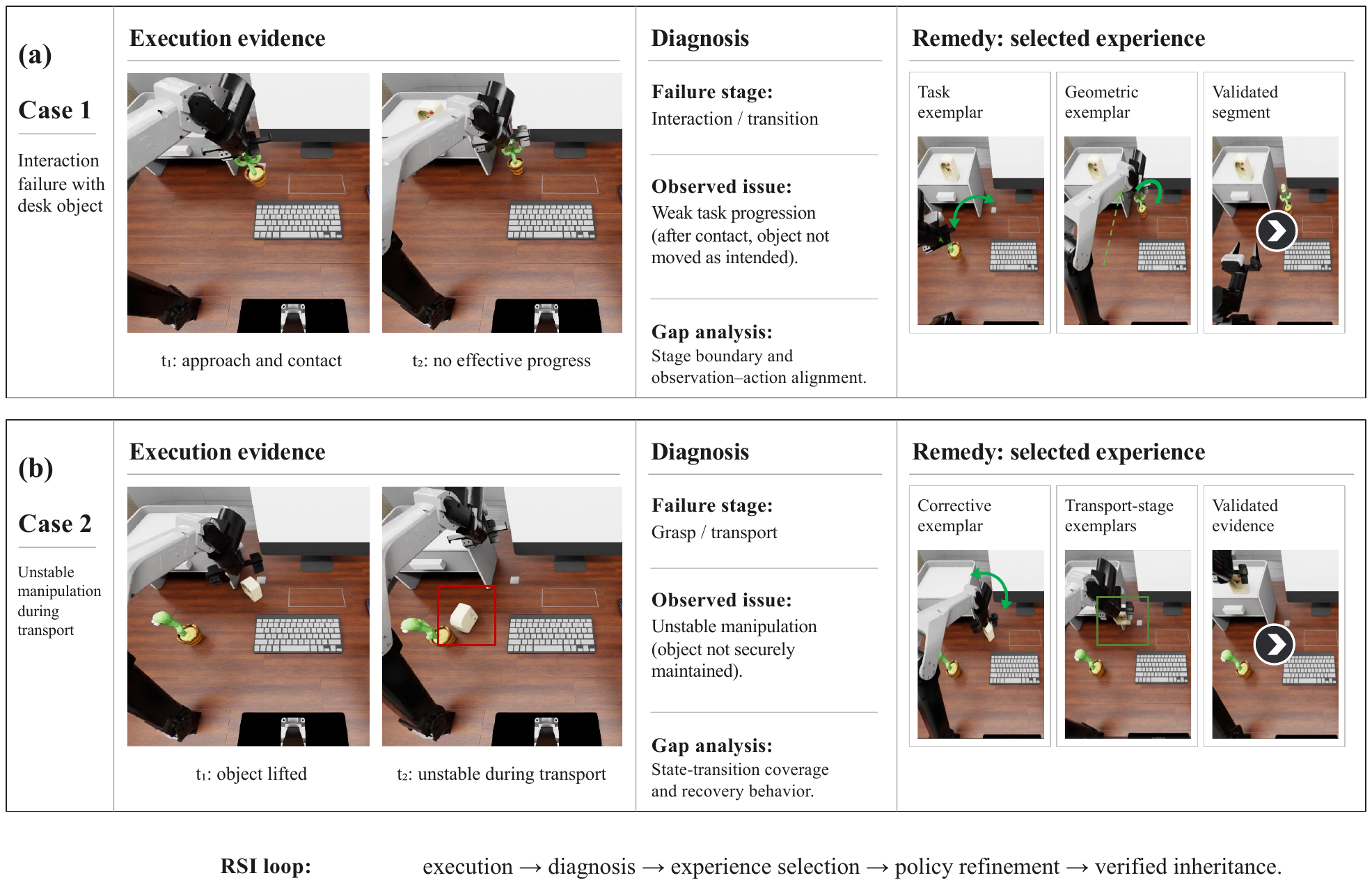}
\caption{Diagnosis and remedy across two representative manipulation failure modes. (a) Contact-to-transition failure on a desk object, where physical contact fails to advance the task stage. (b) In-hand transport instability, where an object is lifted successfully but slips during transport. In both cases, Diagnosis traces execution evidence to the specific failing stage, conditioning the selection of validated corrective experience in Remedy.}
\label{fig:execution_cases}
\end{figure}

Figure~\ref{fig:execution_cases} illustrates this diagnostic-to-remedy linkage on two representative cases. In Case~(a), initial contact with a desk object fails to trigger the intended task progression, localizing the limitation to the boundary between contact establishment and stage transition. In Case~(b), the object is grasped and lifted but loses stability during transport, localizing the limitation to grasp maintenance and in-hand state transition. Case~(b) also exemplifies how the capability frontier shifts across updates: the transport instability becomes observable only after the policy has acquired a reliable lifting capability.

\section{Remedy: Making Failure Teachable}
\label{sec:remedy}
While Diagnosis identifies \emph{where} a policy fails, failed rollouts alone do not specify what the robot should do instead. The Remedy stage bridges this gap by curating validated, stage-targeted experience conditioned on $g_k$ and converting it into a policy update.

\subsection{Unified Episode representation}
Retrieving experience that directly addresses an improvement objective $g_k$ requires a data representation that is jointly queryable by task, stage, and outcome. We standardize all robot experience into a unified, source-aware \emph{Episode} schema that serves as the common unit for representing, validating, and releasing trajectories. Each Episode contains the task identity, language instruction, time-synchronized observation--state--action tuples, source provenance, task-stage boundaries, terminal outcome flags, quality status, and dataset version. Aligning observations, states, actions, and stage transitions along a shared time axis enables the same unit to support diagnostic analysis, stage-level experience selection, and offline policy optimization.

\subsection{Complementary experience sources}
Because a failed rollout lacks the optimal counterfactual action, no single form of experience can supply all that is missing. A geometrically feasible path does not guarantee semantic instruction completion, whereas clean expert demonstrations rarely cover recovery from the specific off-nominal states visited by an imperfect policy. We therefore organize the experience base $\mathcal{B}_k$ around three complementary sources: (i) \emph{task exemplars}, which express instruction-conditioned behavior and feasible multi-stage action sequences; (ii) \emph{geometric exemplars}, which encode target-conditioned spatial structure; and (iii) \emph{on-policy execution evidence}, which captures the actual state distribution induced by $\pi_k$, including successful progressions, incomplete actions, and near-frontier recovery behaviors. Incorporating on-policy execution evidence is vital because offline updates are most effective when supported by experience near the states the policy actually visits \cite{dagger,xiao}.

\subsection{Automated quality validation}
Because offline learning optimizes directly against its dataset, even a small fraction of misaligned Episodes can teach erroneous behavior. Raw execution rollouts require particular care: failed trajectories are indispensable as diagnostic evidence, yet unsuitable as direct action targets. Our automated validation operator $\mathcal{F}$ resolves this tension through a five-stage pipeline: normalization, temporal alignment, task-stage association, consistency assessment, and versioned release. Normalization standardizes observation and state fields; temporal alignment verifies correspondence between observations and control actions; task-stage association maps each trajectory segment to its manipulation phase; and consistency assessment checks coherence among the language instruction, scene state, robot configuration, and action sequence. Incomplete or unstable rollouts are flagged and retained strictly for capability-gap diagnosis ($\mathcal{A}$), whereas only validated, coherent segments join the experience base $\mathcal{B}_{k+1}$ as actionable supervision.

\subsection{RSI-driven experience selection}
As the experience base $\mathcal{B}_{k+1}$ grows across generations, trajectories relevant to the active capability frontier occupy only a small fraction of the repository, so uniform sampling would dilute the corrective signal needed at generation $k$. Instead, our curation operator $\mathcal{C}$ actively selects a targeted subset $\mathcal{D}_{k+1}=\mathcal{C}(\mathcal{B}_{k+1},g_k)$ guided by $g_k$. Specifically, $\mathcal{C}$ retrieves validated Episodes sharing the bottleneck task stage, object-interaction pattern, or state-transition boundary identified in $g_k$, prioritizing trajectories that demonstrate successful progression across the diagnosed limitation. The curated set $\mathcal{D}_{k+1}$ is versioned and linked to the objective $g_k$ and reference policy $\pi_k$, ensuring that every policy update is traceable to the failure mode that motivated it.

\subsection{Objective-conditioned offline reinforcement learning}
To turn the curated dataset $\mathcal{D}_{k+1}$ into improved closed-loop behavior, the update must evaluate actions by their long-horizon consequences across stages \cite{suttonbarto} while learning strictly from a fixed offline experience distribution \cite{levine}. We formulate each remedy update $\mathcal{U}_{\mathrm{off}}$ as an objective-conditioned offline reinforcement learning problem over $\mathcal{D}_{k+1}$, where the improvement objective $g_k$ acts as a goal in the sense of universal value functions \cite{uvfa}. Specifically, the update seeks a policy that maximizes objective-conditioned value while remaining anchored to the inherited reference policy $\pi_k$:
\begin{equation}
\pi'_k=\arg\max_{\pi}\;\mathbb{E}_{s\sim\mathcal{D}_{k+1},\,a\sim\pi(\cdot\mid s,g_k)}\!\left[Q(s,a\mid g_k)\right]-\lambda\,\mathrm{D}(\pi,\pi_k),
\label{eq:offline}
\end{equation}
where $Q(s,a\mid g_k)$ is the learned value estimate on $\mathcal{D}_{k+1}$, $\mathrm{D}(\pi,\pi_k)$ measures divergence from the reference policy, and $\lambda>0$ controls anchor strength. While this formulation follows behavior-regularized and conservative offline reinforcement learning \cite{brac,cql,iql} to mitigate out-of-distribution value extrapolation, the divergence anchor serves a crucial second role in our framework: it constrains unnecessary parameter drift on states and tasks outside the current objective $g_k$. Rather than serving as a final deployment checkpoint, the refined candidate $\pi'_k$ is passed to the Retention stage for verification.

\section{Retention: Making Gains Heritable}
\label{sec:retention}
Although Remedy produces an evolved policy $\pi'_k$ with stronger performance on the targeted capability frontier, updating shared weights inevitably risks degrading previously mastered skills \cite{ewc}. Because overall task success is a product of stage pass rates, even small cross-task regressions can erase net progress. The Retention stage governs what the next generation actually inherits, separating a compounding loop from a sequence of isolated fine-tuning runs.

\subsection{Reference-state initialization}
Capability inheritance requires an explicit predecessor against which behavioral changes can be measured. The framework therefore maintains a shared reference policy $\pi_k$ that can be evaluated as a complete controller at the start of each generation. Compatible policy checkpoints provide its parameter-space basis—with compatibility established by matching model architectures, parameter representations, and execution interfaces—and are integrated via model-soup weight averaging and interference-aware merging \cite{soups,ties}. The resulting reference state gives offline learning ($\mathcal{U}_{\mathrm{off}}$) a coherent initialization that preserves the capabilities verified prior to generation $k$.

\subsection{Candidate consolidation}
Directly adopting the evolved endpoint $\pi'_k$ rarely achieves the optimal balance between new frontier gains and existing skills: full adoption ($\pi'_k$) risks cross-task regression, whereas full rejection ($\pi_k$) discards the newly acquired capability. Consolidation exposes the Pareto frontier between these extremes by constructing candidates along the linear weight-space trajectory from the inherited reference parameters $\theta_k$ to the evolved parameters $\theta'_k$:
\begin{equation}
\theta_k^{(i)}=(1-\alpha_i)\,\theta_k+\alpha_i\,\theta'_k,\qquad \alpha_i\in[0,1],
\label{eq:consolidation}
\end{equation}
where the consolidation coefficient $\alpha_i$ sets how strongly each candidate $\pi_k^{(i)}$ incorporates the update. Searching this linear path is grounded in linear mode connectivity \cite{lmc}: because $\theta_k$ and $\theta'_k$ share a common ancestor, weight-space interpolation retains functional coherence across the loss landscape. Moreover, because interpolation acts directly on weights rather than ensembling action distributions, each candidate $\pi_k^{(i)}$ remains a single standalone policy with zero additional inference overhead.

\subsection{Evaluation-guided promotion}
Because parameter-space proximity alone cannot guarantee skill survival in contact-rich control, retention must be adjudicated on closed-loop behavior across the full task set. Each consolidated candidate $\pi_k^{(i)}$ is evaluated under a standardized RoboDojo-Sim protocol ($\mathcal{V}_{k+1}$) using Overall Score, Success Rate, and task-completion validity, weighing frontier gains on targeted tasks against any loss on others. The selection operator $\mathcal{S}$ promotes the best-performing candidate to $\pi_{k+1}$ and excludes all unpromoted variants.

Promotion is the point at which the loop compounds: the promoted state $\pi_{k+1}$ becomes the inherited reference policy of generation $k+1$, which then executes $\pi_{k+1}$ to diagnose the newly exposed frontier. Model identity, reference lineage, experience version, and evaluation logs are archived together, endowing the policy with a traceable evolutionary history.

\section{Experimental Evaluation}
\subsection{Why RoboDojo-Sim}
Validating whether a recursive framework compounds calls for a broad benchmark where narrow overfitting or cross-task forgetting is immediately penalized in aggregate metrics. We evaluate on \textbf{RoboDojo-Sim}, part of the unified RoboDojo benchmark for generalist robot manipulation \cite{robodojo}. Its 42 tasks span five capability dimensions—\emph{generalization}, \emph{memory}, \emph{long-horizon execution}, \emph{precision}, and \emph{open task specification}—ensuring that a fix that breaks another skill lowers the aggregate result. Overall Score captures continuous stage-level task progress, while Success Rate (SR) measures binary full-task completion. Reading the two metrics together separates policies that merely advance through early stages from those that cross the final frontier to finish the task, matching the diagnostic interpretation emphasized by the benchmark authors \cite{robodojo}. A shared protocol and public leaderboard via XPolicyLab place all policies on equal terms \cite{xpolicylab}, and the benchmark is highly demanding: the highest Success Rate among the thirteen public baselines in Table~\ref{tab:leaderboard} is 14.88\%.

\subsection{Aggregate performance and multi-seed stability}
Both the promoted policy checkpoint (\textbf{CSU-AI-07}) and the benchmark configuration are fixed before evaluation, and performance is reported as the arithmetic mean over three independent evaluation seeds. As shown in Table~\ref{tab:seed_results}, CSU-AI-07 achieves a three-seed mean Overall Score of \textbf{26.15} and a mean Success Rate of \textbf{20.70\%}, with seed-level results spanning a narrow range of 1.07 points and 0.74 percentage points. Notably, while Seed~0 was used for candidate selection during iteration, held-out evaluations on Seeds~1 and~2 remain closely aligned with Seed~0, confirming that evaluation-guided promotion yields stable cross-seed generalization.

\begin{table}[H]
\centering\small
\caption{Full-benchmark RoboDojo-Sim evaluation results for CSU-AI-07 across three independent evaluation seeds.}
\begin{tabularx}{\textwidth}{Y >{\centering\arraybackslash}p{0.24\textwidth} >{\centering\arraybackslash}p{0.24\textwidth}}
\toprule
\textbf{Evaluation} & \textbf{Overall Score $\uparrow$} & \textbf{Success Rate (\%) $\uparrow$}\\
\midrule
Seed 0 & 26.53 & 21.05\\
Seed 1 & 25.46 & 20.31\\
Seed 2 & 26.45 & 20.75\\
\rowcolor{paleBlue}\textbf{Three-seed mean} & \textbf{26.15} & \textbf{20.70}\\
\bottomrule
\end{tabularx}
\label{tab:seed_results}
\end{table}

\begin{table}[t]
\centering\small
\caption{Overall RoboDojo-Sim leaderboard comparing CSU-AI-07 against thirteen public generalist policies. The benchmark protocol follows RoboDojo-Sim \cite{robodojo}.}
\begin{tabularx}{\textwidth}{Y >{\centering\arraybackslash}p{0.22\textwidth} >{\centering\arraybackslash}p{0.25\textwidth}}
\toprule
\textbf{Policy} & \textbf{Overall Score $\uparrow$} & \textbf{Success Rate (\%) $\uparrow$}\\
\midrule
\rowcolor{paleBlue}\textbf{CSU-AI-07} & \textbf{26.15} & \textbf{20.70}\\
GalaxeaVLA (G0.5)~\cite{g05} & 20.23 & 14.88\\
Xiaomi-Robotics-1~\cite{xiaomirobotics1} & 20.07 & 13.93\\
Hy-Embodied-0.5-VLA~\cite{hyvla} & 13.07 & 8.80\\
SimpleMemVLA~\cite{simplememvla} & 12.58 & 9.27\\
Spatial Forcing~\cite{spatialforcing} & 12.38 & 8.04\\
$\pi_{0.5}$~\cite{pi05} & 11.44 & 6.93\\
InternVLA-A1.5~\cite{internvla15} & 11.15 & 7.14\\
VLAct~\cite{vlact} & 10.65 & 7.58\\
X-VLA~\cite{xvla} & 10.13 & 6.52\\
X-WAM~\cite{xwam} & 7.69 & 3.83\\
StarVLA~\cite{starvla} & 6.40 & 3.24\\
$\pi_0$~\cite{pi0} & 3.48 & 1.53\\
GR00T-N1.7~\cite{grootn17} & 2.85 & 1.31\\
\bottomrule
\end{tabularx}
\label{tab:leaderboard}
\end{table}

\subsection{Leaderboard comparison}
Table~\ref{tab:leaderboard} presents the overall RoboDojo-Sim leaderboard, where CSU-AI-07 ranks \textbf{first} on both metrics. It surpasses the next-best Overall Score (20.23) by \textbf{5.92 points} and the next-best Success Rate (14.88\%) by \textbf{5.82 percentage points}—relative margins of \textbf{29.3\%} and \textbf{39.1\%}, respectively. Compared with $\pi_{0.5}$ (6.93\% SR), a widely adopted open generalist policy, CSU-AI-07 achieves nearly three times the task completion rate. At the same time, a 20.70\% Success Rate leaves substantial room for further progress across the 42 tasks; within our framework, these remaining failed trials constitute the empirical evidence that subsequent generations will diagnose and consolidate into inherited capability.

\section{Discussion}
Compounding Physical RSI reframes generalist robot policy development: the central question shifts from how much static experience a policy consumes to how effectively each generation converts its own closed-loop failures into capabilities that its successors inherit. Crucially, the three stages of our framework are mutually indispensable in this loop: without stage-level \emph{Diagnosis}, policy updates lack directional focus at the capability frontier; without targeted \emph{Remedy}, localized failure modes do not translate into corrective supervision; and without evaluation-guided \emph{Retention}, unconstrained updates risk silently erasing the accumulated gains of earlier generations.

The present scope also defines clear next steps. First, while our benchmark results evaluate the integrated system across 42 tasks, fine-grained ablations isolating the individual contributions of diagnosis, experience selection, and weight consolidation will further clarify the mechanics of compounding. Second, compounding here operates on policy capability while the improvement operators ($\mathcal{A}, \mathcal{C}, \mathcal{U}_{\mathrm{off}}$) remain fixed; extending RSI to evolve the improvement procedure itself is the higher bar set by recent formulations of Physical RSI \cite{whoimproves,towardrsi}. Finally, carrying this closed-loop framework from physics simulation to physical hardware—where each trial is far costlier—will make verified capability inheritance even more valuable.

\section{Conclusion}
Because behavior beyond the current capability frontier remains unobserved during execution, a generalist manipulation policy learns what it needs next by executing, failing, and improving. This report has argued that the central challenge of Physical RSI is ensuring that such iterative improvements compound across generations. \textbf{Compounding Physical RSI} addresses this challenge by diagnosing failure at the granularity of task stages, remedying it via objective-conditioned offline reinforcement learning over curated experience, and retaining verified gains through evaluation-guided weight consolidation. The resulting policy, \textbf{CSU-AI-07}, ranks first on the overall RoboDojo-Sim leaderboard across 42 tasks with a three-seed mean Overall Score of 26.15 and Success Rate of 20.70\%, demonstrating that execution failure can be systematically transformed into inherited capability.

\end{document}